\documentclass{article}

\PassOptionsToPackage{numbers, compress}{natbib}

\usepackage[final]{nips_2018}

\usepackage[utf8]{inputenc} 
\usepackage[T1]{fontenc}    
\usepackage{hyperref}       
\usepackage{url}            
\usepackage{booktabs}       
\usepackage{amsfonts}       
\usepackage{microtype}      
\usepackage{graphicx}
\usepackage{amsmath,amssymb}
\usepackage{tikz} 
\usepackage{mathtools}
\usepackage{subcaption}
\usepackage{float}
\DeclareMathOperator{\E}{\mathbb{E}}

\title{Deep Generative Model with Beta Bernoulli Process \\for Modeling and Learning Confounding Factors}

\author{Prashnna K. Gyawali,$^{1}$ Cameron Knight,$^{1}$ Sandesh Ghimire,$^{1}$ B. Milan Horacek,$^2$ \\* \textbf{John L. Sapp},$^2$ \textbf{Linwei Wang}$^{1}$\\
$^{1}$ Rochester Institute of Technology, Rochester, USA\\
$^{2}$ Dalhousie University, Halifax, NS, Canada\\
\{\texttt{pkg2182@rit.edu}\}
}

\begin{document}

\maketitle

\begin{abstract}
While deep representation learning has become increasingly capable of separating task-relevant representations from other confounding factors in the data, two significant challenges remain. First, there is often an unknown and potentially infinite number of confounding factors coinciding in the data. Second, not all of these factors are readily observable. In this paper, we present a deep conditional generative model that learns to disentangle a task-relevant representation from an unknown number of confounding factors that may grow infinitely. This is achieved by marrying the representational power of deep generative models with Bayesian non-parametric factor models. We tested the presented model in a clinical ECG dataset with significant inter-subject variations and augmented with signal artifacts. The empirical results highlighted the ability of the presented model to grow with the complexity of the data and identify the absence or presence of unobserved confounding factors. 
\end{abstract}

\section{Introduction}
Confounding factors are inherent in most data analyses, especially those of clinical data \cite{skelly2012assessing}. 
While recent developments in deep representation learning have increased our ability to separate task-relevant representations of the data from other factors of variation \cite{bengio2013representation}, existing works are yet able to address two important challenges. First, in data collected in a realistic setting, there often coincides an unknown and sometimes infinite number of confounding factors. Examples in clinical data include patient demographics (such as sex and age), disease subgroups, data acquisition modalities, and data artifacts and noises. Second, while some confounding factors are easily observable (such as age and sex), others are not. Examples of the latter include unknown data artifacts/noises, or inter-subject variability in physiological factors such as organ anatomy. 

The Beta Bernoulli process also referred to as an Indian Buffet Process (IBP), is a popular Bayesian nonparametric latent feature model able to represent observations with infinitely many features \cite{griffiths2011indian}. Few initial works \cite{singhstructured, chatzis2014indian} have been presented to use IBP as a prior within VAE. These initial works, however, have not considered 
the context of disentangling or learning confounding factors, 
nor have in-depth analyses been carried out 
regarding how an IBP prior may improve the ability of the VAE 
to grow with the complexity of the data 
or identify the absence or presence of latent features within the data. 
In this paper, we introduce a deep conditional generative model that learns to disentangle a task-relevant representation from an unknown number of confounding factors that may grow infinitely. This is achieved by using a supervised deterministic encoder to learn task-related representation, along with a probabilistic encoder with an IBP prior to model the unknown number of unobservable confounding factors. We term this model as conditional IBP-VAE (cIBP-VAE).

\section{Model}
\subsection{Background: Beta-Bernoulli Process}
Beta-Bernoulli Process, also referred to as the IBP \cite{griffiths2011indian}, is a stochastic process which defines a probability distribution over sparse binary matrices indicating the feature activation for $K$ features. 
The infinite binary sparse matrices is the representation of latent feature allocations $Z \in \{0,1\}^{\{NXK^+\}}$, where $z_{n,k}$ is 1 if feature $k$ is active for the $n^{th}$ sample and 0 otherwise. 
For practical implementations, stick-breaking construction \cite{teh2007stick} is considered where the samples are 
drawn as $\nu_k$ $\sim$ Beta($\alpha, \beta$), $z_{n,k}$ $\sim$ Bernoulli($\pi_k$) and $\pi_k$ = $\prod_{i=1}^{k} \nu_i$.
For brevity, the whole process is written as $Z, \nu \sim$ IBP($\alpha$) with $\alpha$ representing expected number of features of each data point. 


\subsection{Conditional generative model}
In this work, we introduce a conditional probabilistic model admitting two different sources of variations: the task-related representation $y_{t}$,   
and the confounding representation $y_{c}$. 
Considering factor $y_{t}$ as an observed variable, the model can be represented as $X$ $\sim$ $p_{\theta}(X|y_{c},y_{t})$ where the latent variable $y_{c}$ follows a prior distribution $p(y_{c})$. 
To model an unbounded number of unobserved confounders, 
we model $p(y_{c})$ with an IBP prior and arrive at our generative model as: 
\begin{equation}
\begin{aligned}
  \label{eq:genModel}
  Z, \nu \sim IBP(\alpha); A_{n} \sim \mathcal{N}(0, I);
y_{c} = Z \odot A; 
X \sim p_{\theta}(X|Z \odot A,y_{t})
\end{aligned}
\end{equation}
where $\odot$ is the element-wise matrix multiplication operator, known as the Hadamard product. The multiplication with the discrete-variable $Z$ essentially allows the model to infer which latent features captured by $A_{nk}, k \in \{1, .. K \rightarrow \infty\}$ is active for the observed data $X$. 
The likelihood function $p_{\theta}(X|Z \odot A,y_{t})$ is defined by neural networks parameterized by $\theta$. 

\subsection{Inference}
To infer latent variables $Z, A$ and $\nu$, we use variational inference where we propose a variational posterior $q_{\phi_{1}}(Z, A, \nu|X)$ as an approximation for the true posterior $p(Z, A, \nu|X)$.
The decomposition of our variational posterior is performed as
$q_{\phi_{1}}(Z, A, \nu|X) = \prod_{k=1}^{K} q(\nu_{k}) \prod_{n=1}^{N} q(z_{n,k}|\nu_{k}, x_{n})q(A_{n}|x_{n})$
where $q$($\nu_{k}$) = Beta($\nu_{k}|a_{k}$, $b_{k}$), 
$q(z_{n,k}|\nu_{k}, x_{n})$ = 
Bernoulli$(z_{n,k}|\pi_{k}, d(x_{n}))$, 
$\pi_{k}$ = $\prod_{i=1}^{k} \nu_{i}$, 
$q(A_{n}|x_{n})$ = $\mathcal{N}(A_{n}|\mu(x_{n}), \sigma^{2}(x_{n}))$, 
$N$ represents the number of samples, 
and $K$ represents the truncation parameter as an approximation of infinite feature allocation.  
Here, $\nu_{k}$ is the global variable shared between data points, and $z_{n,k}$ and $A_n$ are the local variables. $a_{k}$ and $b_{k}$ are the parameters, and $d(x_{n})$, $\mu(x_{n})$ and $\sigma^{2}(x_{n})$ are encoder networks parameterized by $\phi_{1}$.

To include the  
dependency between conjugate exponential families within the Beta-Bernoulli process, instead of mean-field approximation that makes strong assumptions regarding the independence of the variables, 
we adopt the structured stochastic variational inference (SSVI) 
\cite{hoffman2015structured}. The resulting objective is to maximize the following evidence lower bound (ELBO):
\begin{equation}
\begin{aligned}
  \label{eq:ELBO1}
\mathcal{L} & = - KL(q(\nu_k)||p(\nu_k)) + \sum_{n=1}^{N} \bigg(\E_{q}[\log{p}(x_{n}|Z_{n}, A_{n}, y_{tn})] \\
& - KL(q(Z_{n}|\nu, x_{n})||p(Z_{n}|\nu)) -KL(q(A_{n}| x_{n})||p(A_{n}))\bigg)
\end{aligned}
\end{equation}
Here, we approximate the expectations by first taking a global sample $\nu_{k} \sim q(\nu_k)$, and then sample from $Z_{n} \sim q(Z_{n}|\nu_{k}, x_{n})$ through 
the encoder network. 
This objective function can be interpreted as minimizing a reconstruction error in the second term, 
along with minimizing the KL divergence between the variational posterior approximation and the corresponding priors on all the other terms. 

\subsection{Summary of the model}
The presented generative model is conditioned on 
the task-representation 
$y_{t}$ that is encoded from data $X$ 
through a deterministic encoder $f_{y_{t}}(X)$. 
Due to the use of a deterministic encoder, however, all sources of stochasticity in $y_{t}$ are only from the data distribution. 
To encourage the model to learn discriminative representation in $y_{t}$, 
we extended the unsupervised objective $\mathcal{L}$ in equation (\ref{eq:ELBO1}) with a supervised classification loss: 
\begin{equation}
\begin{aligned}
  \label{eq:finalObj}
\mathcal{L}^{\gamma} =  \mathcal{L} + \zeta \cdot \E_{p(X,y_{t})} [-\log{q_{\phi_{2}}(y_{t}|X)}]
\end{aligned}
\end{equation}
where $\zeta$ controls the relative weight between the generative and discriminative learning. $q_{\phi_{2}}(y_{t}|X)$ is the label predictive distribution \cite{kingma2014semi} approximated by $f_{y_{t}}$(X) parameterized by $\phi_2$.

\section{Experiments}
\vspace{-0.2cm}
We compared the presented cIBP-VAE with c-VAE and other baseline models on a real clinical ECG dataset with significant inter-subject variations and augmented with signal artifacts.
We also augmented the MNIST dataset with colored digits to test the ability of the presented cIBP-VAE. The results are provided in the Appendix A. 
\subsection{Study cohort}
\vspace{-0.1cm}
\label{realData}
A large pace-mapping ECG dataset is collected from 39 scar-related ventricular arrhythmia patients during invasive pace mapping procedures. 
Following pre-processing, as described in ~\cite{chen2017disentangling}, 
we obtained a dataset of 16848 unique 12-lead ECG beats with input size of 1200 (12*100), along with the labeled origins of ventricular activation (pacing sites).  

The clinical ECG dataset comprises an infinite number of
confounding factors 
due to inter-subject variations in 
heart anatomy,
thorax anatomy, 
and pathological remodeling, 
all unobservable. 
To test the ability of the presented cIBP-VAE to 
grow with the complexity of the data, 
we further
augmented this dataset by an artifact 
(of size 10 for each lead)
-- in the form of an artificial pacing stimulus 
-- to $\sim$50\% of all ECG data, selected randomly. 
For the rest, we augment with all 0's of the same size, representing no stimulus.
The entire dataset was split into training, validation and test set where no set shared data from the same patient.

\subsection{Growth with data complexity} 
\vspace{-0.1cm}
To examine how the presented cIBP-VAE grows with the complexity of the data, we considered its performance in 1) classification accuracy and 2) confounder disentanglement before and after the artifacts were introduced to the dataset. For the classification
purpose, the pacing sites are transformed into 10 anatomical segments\footnote{Instead of dividing into anatomical segments, the model could also be trained to predict coordinates \cite{gyawali2017automatic}.} following the setup in \cite{yokokawa2012automated}. Table \ref{tab:growth_table} (left) summarizes the localization accuracy of cIBP-VAE in both settings, in comparison to the three models. Note that we did not include the linear classification model after the pacing artifact is introduced because the extraction of QRS-integrals requires the removal of pacing artifacts. The reported results are found to be statistically significant (p-value < 0.03) and suggests an improved ability of the presented cIBP-VAE to grow with the complexity of the data in comparison to the c-VAE and discriminative CNN model. For the disentanglement analysis, since the confounder factors are unobserved in the clinical dataset, we use the anonymized patient ID as a \textit{weak} label of the confounders. We then test, for cIBP-VAE,
to which extent we can use each of the learned latent representations ($y_{t}$ and $y_{c}$) to classify the task label (pacing segments) and confounder label (patient ID).
The results are summarized in Table \ref{tab:growth_table} (right), where, as shown, $y_{t}$ and $y_{c}$ are each more informative about their respective labels than the other, demonstrating successful disentanglement. Also, the introduction of an additional confounding factor has a minimal effect on the ability of the presented cIBP-VAE to disentangle. 

\begin{table}[t]
        \caption{\small{\textbf{(left)} Segment classification accuracy of the presented method versus three alternative models. \textbf{(right)} Classification accuracy when one factor is associated with the label of the other factor.}}
    \begin{subtable}{.5\linewidth}
      \centering
\begin{tabular}[t]{c|c|c}
    \hline
    Model & w/o artifacts & w artifacts \\
    & (in \%) &  (in \%)\\
    \hline
    QRS Int & 47.61 & - \\ 
    CNN & 53.89 & 52. 44 ($\downarrow 1.45\%$) \\ 
    c-VAE & 55.97 & 53.95 ($\downarrow 2.02\%$) \\ 
    cIBP-VAE & 57.53 & 56.97 ($\downarrow 0.56\%$) \\
    \hline 
  \end{tabular} 
    \end{subtable}%
    \begin{subtable}{.5\linewidth}
      \centering
  \begin{tabular}[t]{c|c|c|c|c}
    \hline
    factor & \multicolumn{2}{c}{anatomical segment} & \multicolumn{2}{c}{patient ID}  \\
    & (w/o) & (w) & (w/o) & (w)\\
    \hline
    $y_{t}$ & 57.53 & 56.97 & 22.06 & 23.09 \\ 
    $y_{c}$ & 31.24 & 30.54 & 66.60 & 66.48\\ 
    \hline
    random & \multicolumn{2}{c}{10} & \multicolumn{2}{c}{4.5} \\
    \hline 
  \end{tabular} 
    \end{subtable} 
            \label{tab:growth_table}
\end{table}

\begin{table}[t]
  \caption{Quantitative reconstruction errors in c-VAE versus cIBP-VAE. Column 2 (all signal):  reconstruction errors for the entire signal including artifacts. Columns 3-5 (only artifacts): reconstruction errors for the artifact portion of the signal, including that calculated for all samples (\textit{all}), 
samples with no pacing artifact (\textit{non-stimulus}), 
and samples with pacing artifacts (\textit{stimulus}). }
  \centering
  \begin{tabular}[t]{c|c|c|c|c}
    \hline
    model & all signal &  \multicolumn{3}{c}{only artifacts}  \\
    &  & all & non-stimulus & stimulus\\
    \hline
    c-VAE & 2293.23 & 3.20 & 3.91 & 2.49 \\ 
    cIBP-VAE & 2273.65 & 0.45 & 0.19 & 0.72\\ 
    \hline
  \end{tabular} 
  \label{tab:reconResult}
\end{table}


\begin{figure*}[t]
\caption{\small{[Best viewed in color] \textbf{(a)} Original signal without artificial pacing stimulus, \textbf{(b)} regular reconstruction, \textbf{(c)} signal generated by turning-off only the triggering unit, \textbf{(d)} signal generated by turning off all the active units of $Z$, and \textbf{(e)} signal generated by turning off all the active units except the triggering unit.}}
\begin{center}
  \includegraphics[width=\linewidth]{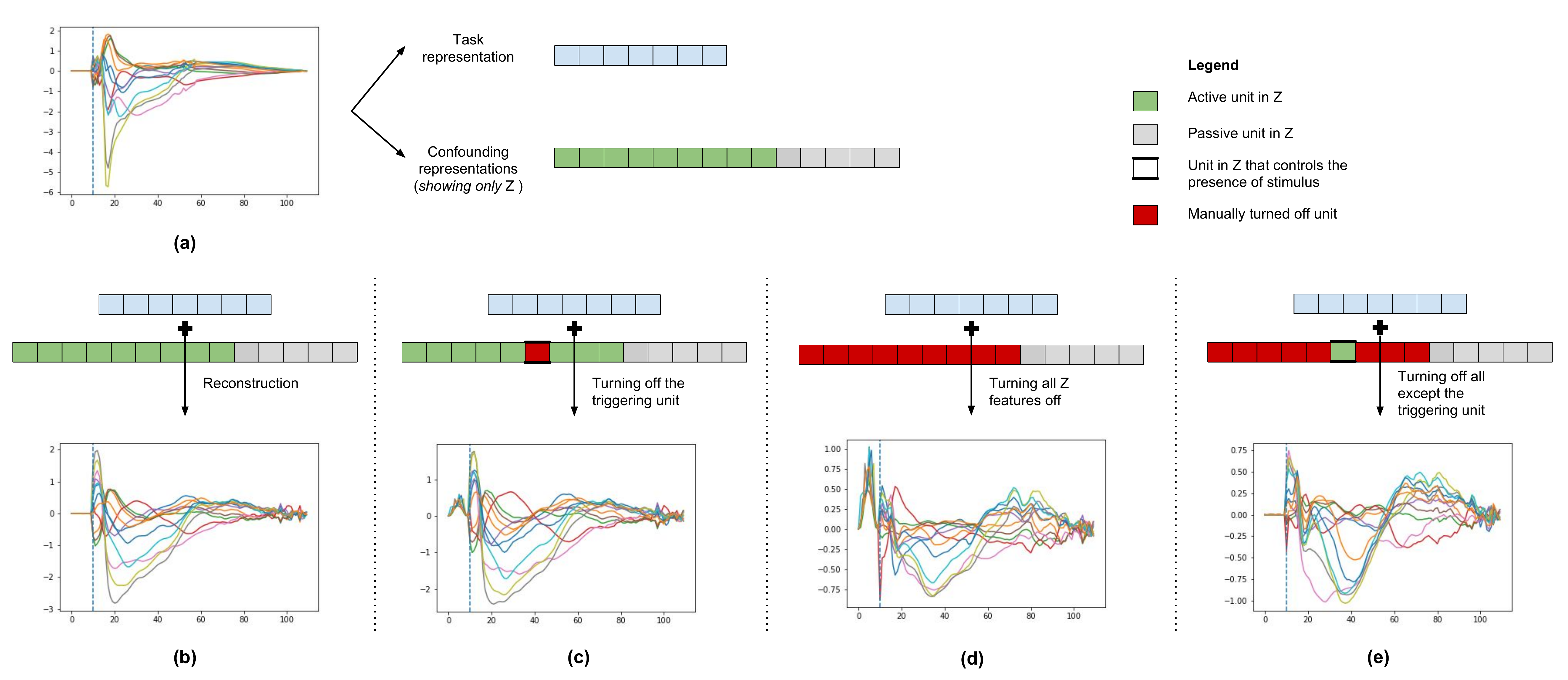}
  \end{center}
\label{fig:trigger.}
\end{figure*}


\subsection{Uncovering latent factors:} 
We then focus on the ability of the two generative models in uncovering 
the binary factor of pacing artifacts in the augmented dataset by considering 1) reconstruction accuracy and 2) identification of binary factor. In terms of reconstruction error, we note that 
while cIBP-VAE and c-VAE 
show a similar overall performance (Table \ref{tab:reconResult}, first column), 
the results may be deceiving because the pacing artifact constitutes only 
a small component of the overall quantitative reconstruction error. 
If we focus on only the portion of the pacing artifacts, 
c-VAE demonstrates a higher error in 
reconstructing data without pacing-artifacts (Table \ref{tab:reconResult}, second column). To understand the ability of the cIBP-VAE to capture the binary confounding factor 
of pacing artifacts, we further analyzed the latent binary features in the cIBP-VAE model. Throughout all test cases, we find that ECG with the pacing artifact is encoded into 24 active binary features, while data without the pacing artifact are encoded into 25 active binary features. In another word, the activation or deactivation of one dimension in the latent binary representations $Z$  is identifying the absence or presence of the confounding artifact in the given data. Through extensive testing of the trained cIBP-VAE, we were able to identify this specific \textit{triggering unit.} An example of an ECG signal without pacing artifact is illustrated in Figure \ref{fig:trigger.}.

\section{Conclusion and Future Work}
The paper presents a deep conditional generative model
for disentangling and learning the unobserved and unbounded number of confounding factors present in the data.
The presented model demonstrated an increased efficacy in learning the task-relevant representations. 
More importantly, the presented model reveal the ability to identify and control the absence or presence of binary confounding factors. 
Future work will focus on analyzing the presented model in a larger variety of health care applications 
where the confounding factors are known to be unbounded. It will also be interesting to further analyze the performance of the model 
in more controlled settings where 
the label information for all or most of the confounding factors is known and can be used for evaluation.

\bibliographystyle{plain}
\bibliography{nips_2018}
\newpage
\appendix
\section{Colored MNIST}
We augment the black-and-white MNIST dataset \cite{lecun1998gradient} 
by adding red, green and blue color to 3/4$^{th}$ of the white characters, resulting in 4 types of colors in the dataset with input size of 2352 (3*28*28).
This adds a binary confounding factor to 
\textit{style} variations 
inherent in the original dataset. 
We focus on the ability of the presented cIBP-VAE to grow with and identify the absence or presence of this binary confounding factor. 

\begin{figure}[H]
\begin{center}
  \includegraphics[width=\linewidth]{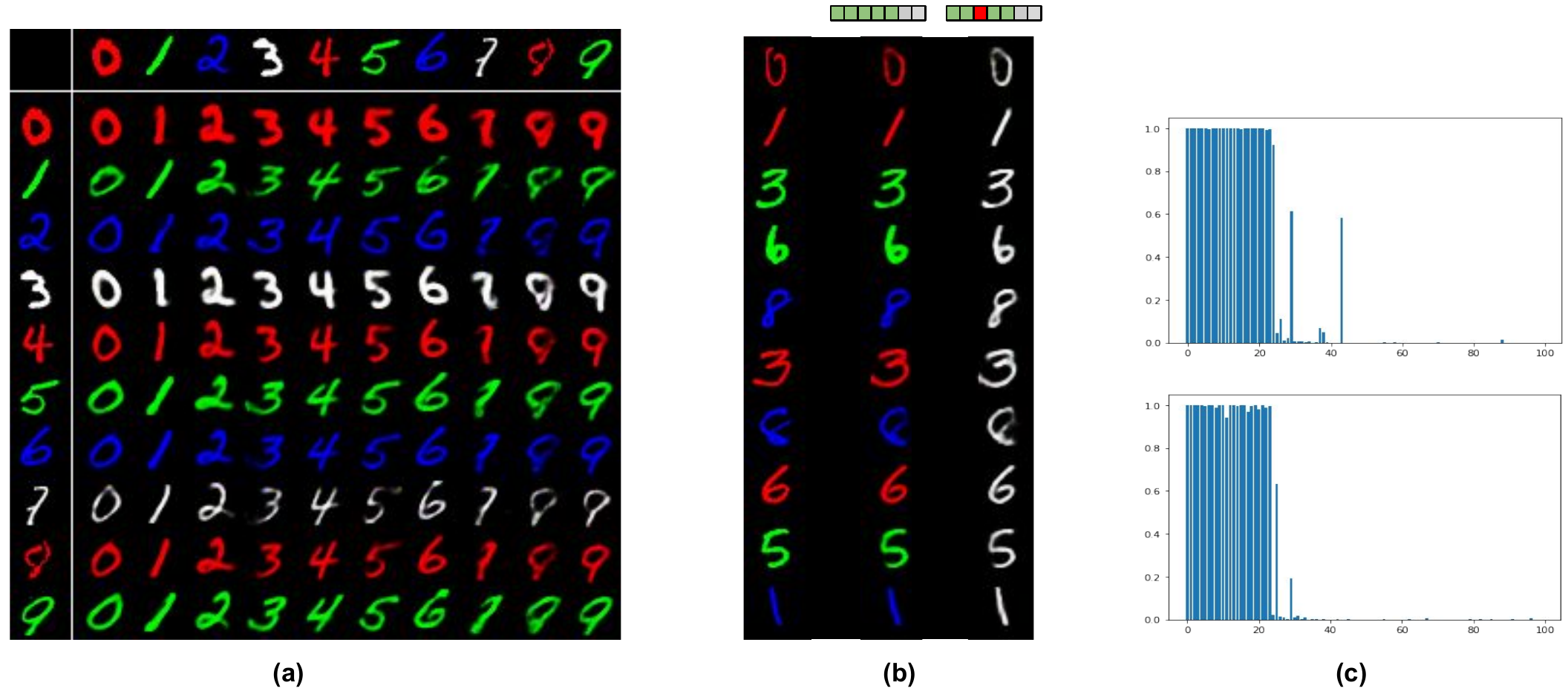}
  \end{center}
\caption{\small{[Best viewed in color] \textbf{(a)} A visualization grid of colored MNIST images by swapping the digit representation from the top-most row with the confounding representation from the left-most column, \emph{i.e.}, hand-writing style and color are transferred from the digits on the leftmost column to each of the ten digits,  \textbf{(b)} Examples of original images (left column), 
reconstructions (middle column), and reconstructions with manually de-activating the triggering units (right column). \textbf{(c)} An example of the bar diagram of the active units in $Z$, showing a higher number of active features for colored image (top) compared to non-colored image (bottom). }}
\label{fig:MNIST_disentanglement}
\end{figure}

Figure \ref{fig:MNIST_disentanglement} shows examples 
of swapping $y_{t}$ (digit representation) 
and $y_{c}$ (confounding representation) from a pair of input images using the trained cIBP-VAE. As shown, 
the confounding representation extracted from one image 
(left most of each row) -- including both the 
style and color information -- can be well transferred to most of the ten digits. 

Within the confounding representation $y_{c}$, throughout all test cases, we found that colored images were encoded with 26 active binary features, while non-colored images(\textit{i.e.}, white characters) were encoded with 24 active binary features (Figure \ref{fig:MNIST_disentanglement} (c)). This suggests that specific units within the binary features $Z$ were responsible for recognizing the absence or presence of color in these images. 
We identified these \textit{triggering units} through extensive testing on the trained network, which can be de-activated to remove colors from a given image. Figure \ref{fig:MNIST_disentanglement} (b) provides such examples where the color of a digit was removed by manually turning off the same triggering units in the network. 

\end{document}